\documentclass[article, 10pt, conference]{ieeeconf}      

\IEEEoverridecommandlockouts                              

\usepackage{spconf}

\usepackage{amssymb}
\usepackage{graphicx,epsfig,setspace,subfig,url,amsmath}
\usepackage{subfig}
\usepackage{longtable}
\usepackage{array}
\usepackage{amsmath}
\usepackage{mathtools}
\usepackage{caption}
\usepackage{multirow}
\usepackage[usenames, dvipsnames]{color}

\renewcommand{\baselinestretch}{0.95} \normalsize

\usepackage{algorithm}
\usepackage[noend]{algpseudocode}

\title{Model Agnostic Saliency for Weakly Supervised Lesion Detection from Breast DCE-MRI 
}

 \name{Gabriel Maicas$^{\dagger}$ \quad Gerard Snaauw$^{\dagger\ddagger}$ \quad Andrew P. Bradley$^{\dagger\dagger}$ \quad   Ian Reid$^{\dagger}$ \quad Gustavo Carneiro$^{\dagger}$ \sthanks{Supported by Australian Research Council through grants DP180103232, CE140100016 and FL130100102.}}
 \address {$^{\dagger}$ Australian Institute for Machine Learning, School of Computer Science, The University of  Adelaide \\
 $^{\ddagger}$Faculty of Applied Sciences, Delft University of Technology \\
 $^{\dagger\dagger}$ Science and Engineering Faculty, Queensland University of Technology }

\renewcommand{\thesection}{\arabic{section}} 
\renewcommand{\thesubsection}{\thesection.\arabic{subsection}}

\begin{document}

\maketitle
\thispagestyle{empty}
\pagestyle{empty}

\begin{abstract}

There is a heated debate on how to interpret the decisions provided by deep learning models (DLM), where the main approaches rely on the visualization of salient regions to interpret the DLM classification process.
However, these approaches generally fail to satisfy three conditions for the problem of lesion detection from medical images: 1) for images with lesions, all salient regions should represent lesions, 2) for images containing no lesions, no salient region should be produced, and 3) lesions are generally small with relatively smooth borders.
We propose a new model-agnostic paradigm to interpret DLM classification decisions supported by a novel definition of saliency that incorporates the conditions above.
Our model-agnostic 1-class saliency detector (MASD) is tested on weakly supervised breast lesion detection from DCE-MRI, achieving state-of-the-art detection accuracy when compared to current visualization methods.

\end{abstract}
\begin{keywords}
saliency, weakly supervised detection, model interpretability, diagnosis explanation, breast lesion localization, breast magnetic resonance imaging.
\end{keywords}

\section{Introduction}
\label{sec:intro}

There is growing debate concerning the interpretation of classifications made by deep learning  models (DLM)~\cite{lipton2016mythos}, particularly in medical diagnosis systems that can directly influence treatment decisions~\cite{goodman2016european}. 
The clinical acceptance of DLMs depends, among other factors, on a reliable explanation of the model outcomes~\cite{caruana2015intelligible}.
A popular approach that can "explain" DLM predictions relies on a salient region detector~\cite{smilkov2017smoothgrad}.
Such weakly supervised DLMs are trained to perform binary classification (negative: no lesion, positive: lesions) and produce salient regions that are assumed to highlight the regions responsible for the positive classification~\cite{smilkov2017smoothgrad}. However, this assumption is unwarranted for two reasons.
For positive classifications, there is no guarantee that salient regions represent lesions, and for negative volumes, salient regions have an unclear meaning.  
We argue that these issues stem from the fact that saliency is poorly defined for weakly supervised DLMs, where the training set contains images with global annotations (i.e image-level labels), but no lesion delineation. 

\begin{figure*}[t]
\begin{center}
\includegraphics[width=1\textwidth]{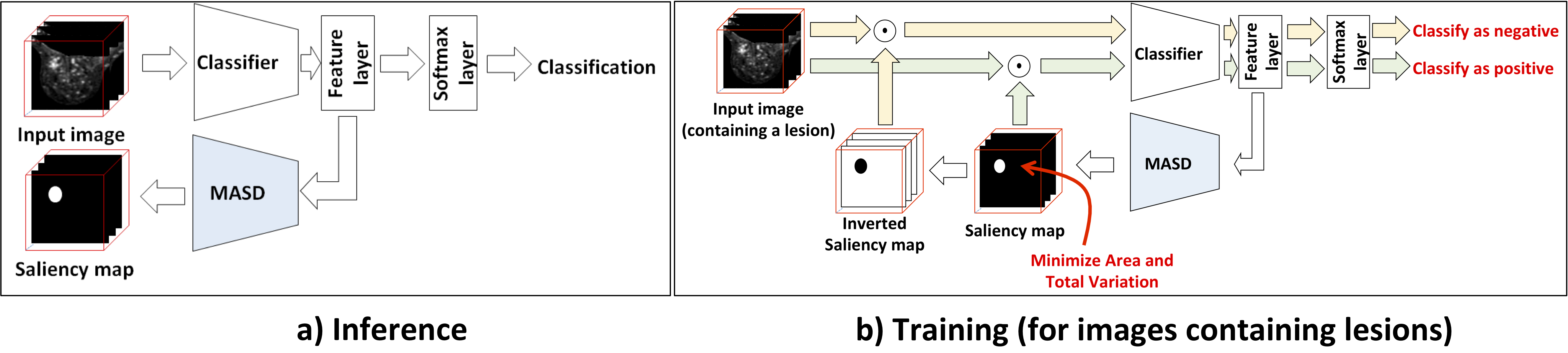} \end{center}
\vspace{-.1in} 
\caption{During inference (a), the weakly trained DLM produces a classification for the presence of a breast lesion -- if the classification is positive, our proposed MASD outputs a saliency map that highlights the regions containing lesions.  The training for volumes containing lesions in (b) is based on a loss function that penalizes: 1) saliency maps that are not smooth and contain large regions, 2) negative classification from the input volume filtered by the saliency map, and 3) positive classifications from the input volume filtered by the inverse of the saliency map.  The training process for volumes that do not contain lesions penalizes the presence of any active region in the saliency map.}
\label{fig:intro}
\end{figure*}

We propose a new paradigm for explaining DLM classifications  supported by a novel saliency definition for the problem of lesion detection. We define saliency as an image region that is responsible for a positive classification as opposed to previous saliency definition that was simply assumed to be active image regions during classification.
Our new model-agnostic 1-class saliency detector (MASD) is explicitly trained to detect lesions from volumes that have been classified by a separate DLM (see Fig.~\ref{fig:intro} - note that MASD is independent of the DLM classifier, which is the reason why we call it model-agnostic).  This goal is achieved by explicitly defining salient regions as follows: 
1) they only appear when the volume is positively classified; 2) they have a small area and a relatively smooth boundary; 3) when used to mask a positively classified volume, it remains positively classified; and 4) when the inverted salient regions are used to mask a positively classified volume, it becomes negatively classified.
The design of MASD is adapted from recent saliency detectors~\cite{dabkowski2017real,fong2017interpretable} to become a 1-class saliency detector by incorporating our saliency definition above. 
We test our approach on two weakly supervised lesion detection problems from breast dynamic contrast-enhanced magnetic resonance volumes (DCE-MRI): 1) (benign and malignant) lesion detection, and 2) malignant lesion detection.
We show that our results are more accurate than the ones produced by the following state-of-the-art (SOTA) saliency detectors: CAM~\cite{zhou2016learning}, a popular approach for weakly supervised detection; Grad-CAM and Guided-GRAD-CAM~\cite{selvaraju2016grad}, which are two extensions that improve upon CAM.

\section{Literature Review}
\label{sec:lit_review}

DCE-MRI is recommended as a complementary imaging modality for screening patients at high-risk for breast cancer~\cite{smith2017cancer}. 
Computer-aided detection methods have been designed~\cite{gubern2015automated,maicas2017deep,mcclymont2014fully,amit2017hybrid} and trained with strong (voxel-wise) annotations to assist radiologists.
As strong annotations are time consuming and noisy, this approach does not scale to the large datasets necessary for deep learning systems.  
An alternative approach is to use the large weakly labeled datasets that are more readily available.
The main challenge behind this approach is the requirement that the classifier should not only classify the case, but also highlight the region containing the lesion for the positive cases.

The medical imaging community has addressed this challenge by exploring saliency detectors~\cite{feng2017discriminative,wang2017chestx}. These approaches are based on highlighting regions of an image that are involved in the classification of each visual class by looking at the activations~\cite{zhou2016learning}.
However, they do not work well when searching for tumors in breast DCE-MRI (given their inconsistency in shape and appearance~\cite{oliver2010review}).  
Other approaches addressed these issues~\cite{wang2017zoom,selvaraju2016grad}, but none of them consider the major weakness of such saliency detection -- the assumption that \textbf{salient regions represent lesions}.  

Dabkowski and Gal~\cite{dabkowski2017real} extended the work by Fong et al.~\cite{fong2017interpretable} by guaranteeing that salient regions represented the visual class associated with the classification by explicitly defining saliency.
They introduced a saliency loss function that finds a saliency mask such that: 1) the classification confidence is not perturbed when the image is masked and 2) the classification confidence is reduced when masked image regions are removed. 
However, there is no definition for tumors and lesions that address the characteristics of saliency in medical images.
We propose a 1-class saliency detector that is able to  interpret the classification of breast DCE-MRI volumes to detect lesions. Our main contribution lies on the explicit definition of saliency based on the definition of a breast lesion that then allows the definition of an appropriate loss function.

\section{Methodology}
\label{sec:methodology}

\subsection{Dataset}
\label{sec:method_data}

The DCE-MRI dataset is defined by ${\cal D} = \left \{ \left ( \mathbf{x}_i, \mathbf{t}_i, \{ \mathbf{s}_i^{(j)} \}_{j = 1}^{S_i} , b_i, y_i \right ) \right \}_{i=1}^{N}$, where $\mathbf{t}_i:\Omega \rightarrow \mathbb R$ represents the T1-weighted volume (with $\Omega$ being the volume lattice), $\mathbf{x}_i:\Omega \rightarrow \mathbb R$ denotes the DCE-MRI first subtraction volume,
the segmentation map
$\mathbf{s}^{(j)}:\Omega \rightarrow \{0,1\}$ is a binary volume indicating the presence or absence of lesion at each voxel for one of patient $i$'s $S_i$ lesions (note that we use this annotation only for testing our approach -- \emph{not for training}), $b_i \in \{ \text{left} , \text{right}  \}$ indicates if this is the left or right breast of the patient, and $y_i \in \mathcal{Y} = \{0,1,2\}$ denotes the breast label ($y_i = 2$: breast contains a malignant lesion, $y_i = 1$ : breast contains at least one benign and no malignant findings, and $y_i = 0$ : no findings). 
We consider two scenarios: 1) \textbf{lesion detection}, where labels $y \in \{1,2\}$ are joined into the positive class, and $y=0$ represents the negative class; and 
2) \textbf{malignant lesion detection}, where labels $y \in \{0,1\}$ are joined into the negative class, and $y=2$ represents the positive class.
We divide ${\cal D}$ in a patient wise manner into training, validation and testing with no overlap between sets.

\subsection{Model Agnostic 1-class Saliency Detector (MASD)}

Our proposed MASD model adapts the saliency detector of Dabkowski and Gal~\cite{dabkowski2017real} to work as a 1-class detector.
The saliency detector in~\cite{dabkowski2017real} uses a encoder-decoder structure with skip connections. The encoder is fixed and trained to produce the classification of the input image, and the decoder generates a mask of the same size of the input and is trained with a loss function that implements the saliency loss described in the last paragraph of Sec.~\ref{sec:lit_review}. In~\cite{dabkowski2017real}, the encoder and decoder are connected through a class selector at the lowest resolution, indicating the visual classes of the salient regions.

\begin{figure}[t]
\begin{center}
\includegraphics[width=.48\textwidth]{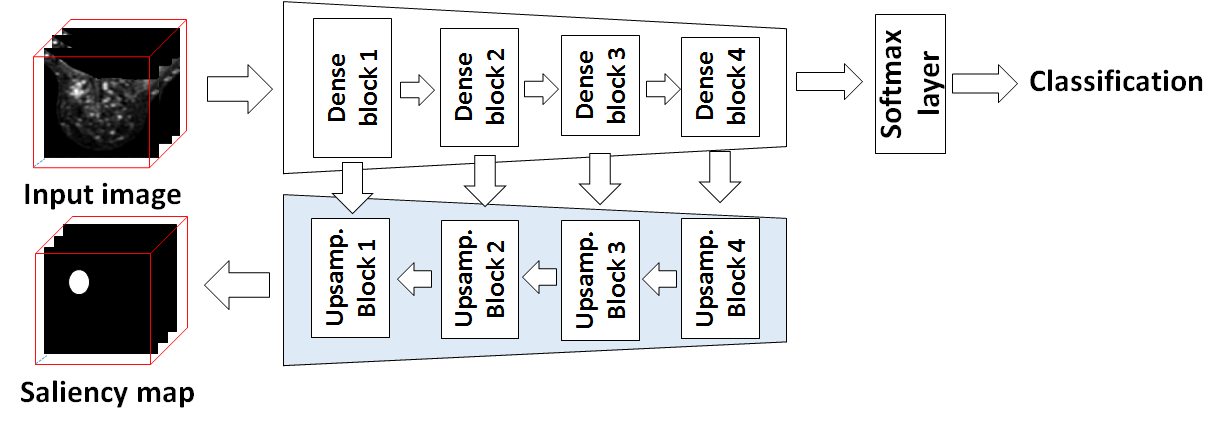} \end{center}
\vspace{-.2in} 
\caption{MASD model diagram.}
\label{fig:MASD}
\end{figure}

Our model (see Fig.~\ref{fig:MASD}) uses a similar structure but with no class selector. The encoder consists of a four-block DenseNet~\cite{huang2017densely} trained for the corresponding weakly supervised classification problem.
The decoder consists of four blocks, with each block comprising a feature map resize, a convolution layer, a batch normalization layer and ReLU activation~\cite{zeiler2014visualizing} -- the decoder outputs the saliency mask $\mathbf{m}:\Omega \rightarrow [0,1]$.
Our contribution lies in the removal of the class selector and modification of the loss function to become a 1-class saliency detector that produces salient regions with the lesion properties defined in Sec~\ref{sec:intro} ($2^{nd}$ paragraph).
The loss function used during \textbf{training} for each sample $i$ is:
\begin{equation}
\begin{split}
\ell_i(\mathbf{m})= &\lambda_1 \ell_{TV}(\mathbf{m}) + \lambda_2 \ell_{A}(\mathbf{m}) - y_i \lambda_3  \ell_{P}(\mathbf{m},\mathbf{x}_i)  \\
&+ y_i \lambda_4 \ell_{D}(1-\mathbf{m},\mathbf{x}_i),
\label{eq:saliencyLoss}
\end{split}
\end{equation}
where $\ell_{TV}(\mathbf{m})$ denotes the total variation of the mask and encourages 
detected regions to be relatively smooth, $\ell_{A}(\mathbf{m})$ computes the area of salient regions and penalizes large regions in the mask,
$\ell_{P}(\mathbf{m},\mathbf{x}_i) = \log P(y=1 | \phi(\mathbf{m},\mathbf{x} ) )$  aims to  maximize the positive classification when the input volume is masked with the saliency mask (i.e. $\phi(\mathbf{m},\mathbf{x})$), and
$\ell_{D}(1-\mathbf{m},\mathbf{x}_i) = P(y=1 | \phi(1-\mathbf{m},\mathbf{x} ) )  $ aims to  minimize the positive classification when the regions of the mask are removed from the volume.  
The loss function~(\ref{eq:saliencyLoss}) is designed for the model to  become a 1-class saliency detector and respond only to a single class (i.e lesions).
It minimizes the number and area of regions in both negative and positive volumes and masks only positive volumes (avoiding the class selector), which are the ones containing salient regions. Note that $y_i = 0$ switches off $\ell_{P}$ and $\ell_{D}$ losses so that volumes classified as negative will not produce any salient regions.

During \textbf{inference}, the saliency mask is generated with a forward pass of the whole architecture, and we threshold the output mask $\mathbf{m}$ and form $\mathbf{m}^{(\tau)}: \Omega \rightarrow \{0,1\}$, such that $\mathbf{m}_{ijk}^{(\tau)} = 1$, if $\mathbf{m}_{ijk} > \tau$, and $\mathbf{m}_{ijk}^{(\tau)} = 0$, otherwise.

\section{Experiments and Results}
\label{sec:Experiments}

The dataset used to assess MASD contains 117 DCE-MRI and T1-weighted volumes (one DCE-MRI and one T1-weighted volume per patient)~\cite{mcclymont2014fully,maicas2017deep}. The training, validation and test set contain 45, 13, and 59 patients~\cite{maicas2017deep} respectively.
The number of lesions for each of the above sets is 57 (38 malignant (m) and 19 benign (b)), 15 (11 m and 4 b), and 69 (46 m and 23 b). 
The T1-weighted volume is used to automatically extract the left and right breasts into separate volumes of $100\times100\times50$ voxels~\cite{maicas2017deep}, and the DCE-MRI volume is used for the classification and lesion detection approaches.

For the MASD model,
the localization is performed only on positively classified samples, defined by a  probability of being positive higher than the equal error rate (EER) of the classifier. Note that EER is computed using the validation set.
We perform experiments on the two problems defined in Sec.~\ref{sec:method_data}: \textbf{lesion detection} and \textbf{malignant lesion detection}. We train a DenseNet as the base classifier for each of the problems.
The parameters, estimated with the validation set using grid search, in (\ref{eq:saliencyLoss}) are 
$\lambda_1 = 0.1$, $\lambda_2 = 2 $, $\lambda_3 = 0.3$, and $\lambda_4 = 2$ for lesion detection and $\lambda_1 = 0.1$, $\lambda_2 = 3$, $\lambda_3 = 1 $, and $\lambda_4 = 2.5$ for malignant lesion detection problem.
We evaluate our methodology in terms of true positive rate (TPR) and the number of false positive detections (FPD) per patient, which are plotted as a free response operating characteristic curve (FROC) (obtained by thresholding the mask produced by MASD at different values in $[0,1]$).
For each of the two problems above, we present results for two different scenarios: 
1) \textbf{All}: the TPRs are computed using the total number of lesions (problem 1: lesion detection) and the total number of malignant lesions (problem 2: malignant lesion detection) in the dataset, and the FPDs per patient rates are computed using the total number of patients with lesions (problem 1), and the number of patients that have malignant lesions (problem 2); and
2) $\mathbf{C +}$: the TPRs and FPDs are computed as above, but we disregard all volumes classified as negative by the DenseNet classifier.
The second scenario above isolates the performance of MASD, which is the main contribution of this paper.
A true positive is considered to be a detection if it has Dice $\geq 0.2$~\cite{maicas2017deep}.
Finally, we provide a comparison with current SOTA weakly supervised region detectors: CAM~\cite{zhou2016learning}, GRAD-CAM~\cite{selvaraju2016grad} and Guided-Grad-CAM~\cite{selvaraju2016grad}.

\begin{figure}[!ht]
\centering
	\begin{tabular}{c}
	\includegraphics[width=0.375\textwidth]{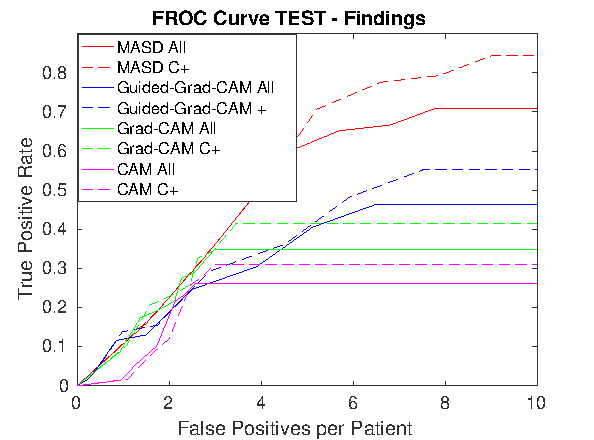} \\
	\end{tabular}
\qquad
	\begin{tabular}{c}
	\includegraphics[width=0.375\textwidth]{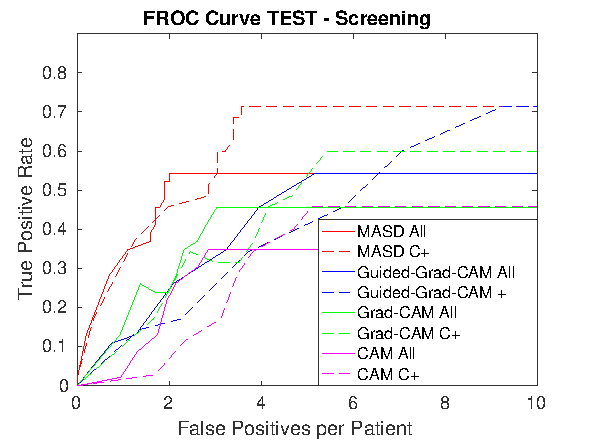} \\
	\end{tabular}
\caption{FROC curves showing TPR vs FPD for \textbf{lesion detection} (Top) and \textbf{malignant lesion detection} (Bottom) for the \textbf{All} (solid curves) and \textbf{C+} (dashed) experiments.}
\label{fig:frocs}
\end{figure}

\begin{figure}[t]
\centering
\subfloat[]{\includegraphics[width = 0.15\textwidth]{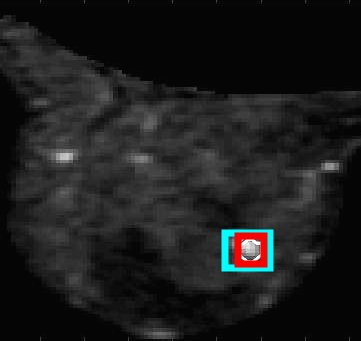}\label{img1}}\hspace{0.07in}
\subfloat[]{\includegraphics[width = 0.15\textwidth]{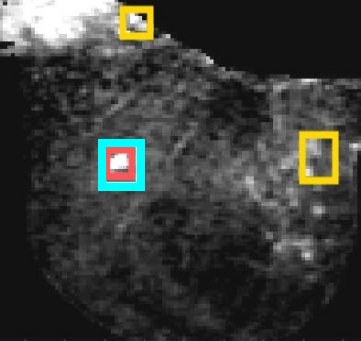}\label{img3}}\hspace{0.07in}
\subfloat[]{\includegraphics[width = 0.15\textwidth]{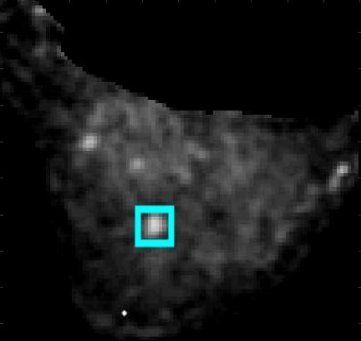}\label{img4}}
\caption{Detections obtained with MASD for \textbf{lesion detection}. Red boxes mark true positive detections, yellow boxes false positive detections, and cyan the ground truth boxes.}
\label{fig:visualDet}
\end{figure}

Figure~\ref{fig:frocs} shows the FROC curves for each of the problems detailed above, and Fig.~\ref{fig:visualDet} shows MASD results for the \textbf{lesion detection} problem in test images.

The results in Fig.~\ref{fig:frocs} show that our proposed MASD approach is more accurate for both problems of lesion and malignant lesion detection than current SOTA saliency visualization methods~\cite{zhou2016learning,selvaraju2016grad}, where only Guided-Grad-CAM~\cite{selvaraju2016grad} achieves a relatively competitive result.  
We believe that the reason behind this Guided-Grad-Cam result lies in the use of guided backpropagation that enables it to capture the fine details of the salient regions, but the mistakes made in low resolution are also carried over, increasing the FPD.
Using the DenseNet classification result to filter out the negatively classified samples, all methods present slightly improved FROC curves, where the TPR curve tends to be consistently higher for the same rates of FPDs per patient.
Compared to strongly supervised lesion detection approaches~\cite{maicas2017deep,mcclymont2014fully}, which show a TPR=0.8 for FPD=3 and TPR=0.9 for FPD=4, the MASD approach still has room for improvement -- for instance, a TPR=0.8 happens only at FPD=8.  We conjecture that more competitive results can be achieved with significantly larger datasets, but the empirical confirmation of this supposition is left for future work.

We tested the influence of each term in the loss function (\ref{eq:saliencyLoss}) and 
we found that the terms that produced the largest variation in our results were $\ell_{D}(.) $ and $\ell_{A}(.)$ because they allowed a significant reduction in the number and size of salient regions. 
The $\ell_{TV}(.)$ and  $\ell_{P}(.)$ terms mainly influenced the lesion detection problem by increasing the FPDs.


 \section{Conclusion}
\label{sec:dic}

We proposed MASD, a model agnostic 1-class saliency detector that can localize lesions in weakly supervised classification problems from breast DCE-MRI. By designing a loss function that explicitly incorporates terms that define a lesion (e.g. size, masked volume classification performance, absence in negative images), we demonstrate that the detected salient regions are more likely to represent the lesions that explain the decision process of deep learning classifiers.  
We believe that explaining the decision process of weakly-supervised classifiers will become a dominating aspect in the field because it is likely that doctors will require an explanation that can justify a DLM classification~\cite{caruana2015intelligible}.

We would like to thank Nvidia for the donation of a TitanXp that supported this work.

\bibliographystyle{IEEEbib}
\bibliography{bibli.bib}

\end{document}